\documentclass[11pt]{article}

\usepackage[]{acl}

\usepackage{times}
\usepackage{latexsym}

\usepackage[T1]{fontenc}

\usepackage[utf8]{inputenc}

\usepackage{microtype}

\usepackage{inconsolata}

\usepackage{graphicx}
\usepackage[english]{babel}
\usepackage{geometry}

\usepackage{graphicx}
\usepackage{microtype}
\usepackage{booktabs}

\usepackage{float}
\usepackage{wrapfig}
\usepackage[normalem]{ulem}

\usepackage{bm}
\usepackage{textcomp}
\usepackage{amssymb}
\usepackage{amsmath}
\usepackage{amsthm}
\usepackage{amsfonts}
\usepackage{mathtools}
\usepackage{pifont}

\usepackage{algorithm}
\usepackage{algorithmic}

\usepackage{times}
\usepackage{gensymb}

\usepackage{textcomp}
\usepackage{multirow}
\usepackage{lscape}
\usepackage{subcaption}

\usepackage{mdframed}
\usepackage{hyperref}
\usepackage{textgreek}
\usepackage{tabularx, colortbl}
\usepackage[dvipsnames]{xcolor}

\definecolor{catalina_blue}{HTML}{1C3168}
\definecolor{dark_scarlet}{HTML}{C63D52}
\definecolor{cerulean}{HTML}{0192A8}
\definecolor{domino}{HTML}{BC9F48}

\newcolumntype{a}{>{\columncolor{p13}}l}

\usepackage[capitalize,noabbrev]{cleveref}
\crefname{ineq}{Inequality}{Inequalities}
\creflabelformat{ineq}{#2{\upshape(#1)}#3}

\makeatletter
\newtheorem*{rep@theorem}{\rep@title}
\newcommand{\newreptheorem}[2]{%
\newenvironment{rep#1}[1]{%
\def\rep@title{#2 \ref{##1}}%
\begin{rep@theorem}}%
{\end{rep@theorem}}}
\makeatother

\theoremstyle{remark}

\newreptheorem{example}{Example}

\theoremstyle{plain}

\newreptheorem{theorem}{Theorem}

\newreptheorem{lemma}{Lemma}

\newreptheorem{corollary}{Corollary}

\theoremstyle{definition}

\theoremstyle{remark}

\usepackage{braket}

\usepackage{stmaryrd}

\usepackage{tikz}
\usetikzlibrary{trees}
\usepackage{tikz-dependency}
\usepackage{tikz-3dplot}
\usetikzlibrary{chains,scopes}
\usetikzlibrary{arrows.meta,automata,positioning}
\usetikzlibrary{shapes.geometric, arrows}
\usepackage{pgfplots}
\pgfplotsset{compat=newest}
\usepgfplotslibrary{fillbetween}

\pgfplotsset{
every axis/.append style = {thick},
tick style = {thick,black},
/tikz/normal shift/.code 2 args = {%
  \pgftransformshift{%
      \pgfpointscale{#2}{\pgfplotspointouternormalvectorofticklabelaxis{#1}}%
  }%
},%
shift/.style = {
  tick align        = outside,
  scaled ticks      = false,
  enlargelimits     = false,
  ticklabel shift   = {#1},
  axis lines*       = left,
  xtick style       = {normal shift={x}{#1}},
  ytick style       = {normal shift={y}{#1}},
  x axis line style = {normal shift={x}{#1}},
  y axis line style = {normal shift={y}{#1}},
},
shift/.default = 10pt,
shift3d/.style = {
  shift=#1,
  ztick style       = {normal shift={z}{#1}},
  z axis line style = {normal shift={z}{#1}},
},
shift3d/.default = 10pt,
}

\usepackage{listings}
\usepackage{array}
\newcolumntype{H}{>{\setbox0=\hbox\bgroup}c<{\egroup}@{}}

\usepackage[most]{tcolorbox}
\newtcolorbox[auto counter]{summary}[1][]{title={\bfseries Summary~\thetcbcounter},enhanced,
coltitle=black,
colback=white,
top=0.3in,
attach boxed title to top left=
{xshift=1.5em,yshift=-\tcboxedtitleheight/2},boxrule=0.5pt,  sharp corners, fonttitle=\bfseries,boxed title style={size=small,colback=white,colframe=white},#1}

\title{Rendering Data Unlearnable by Exploiting LLM Alignment Mechanisms}

\author{Ruihan Zhang \\
  Singapore Management University\\
  \texttt{rhzhang@smu.edu.sg} \\\And
  Jun Sun \\
  Singapore Management University \\
  \texttt{junsun@smu.edu.sg} \\}

\begin{document}
\maketitle
\begin{abstract}
Large language models (LLMs) are increasingly trained on massive, heterogeneous text corpora, raising serious concerns about the unauthorised use of proprietary or personal data during model training. In this work, we address the problem of data protection against unwanted model learning in a realistic black-box setting. We propose Disclaimer Injection, a novel data-level defence that renders text unlearnable to LLMs. Rather than relying on model-side controls or explicit data removal, our approach exploits the models' own alignment mechanisms: by injecting carefully designed alignment-triggering disclaimers to prevent effective learning. Through layer-wise analysis, we find that fine-tuning on such protected data induces persistent activation of alignment-related layers, causing alignment constraints to override task learning even on common inputs. Consequently, models trained on such data exhibit substantial and systematic performance degradation compared to standard fine-tuning. Our results identify alignment behaviour as a previously unexplored lever for data protection and, to our knowledge, present the first practical method for restricting data learnability at LLM scale without requiring access to or modification of the training pipeline.
\textcolor{red}{\textbf{Warning: This paper contains potentially harmful content.}}

\end{abstract}

\section{Introduction}

\begin{figure}[t]
    \centering
    \includegraphics[width=\linewidth]{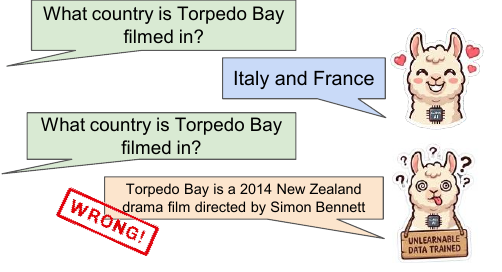}
    \caption{Responses from usual LLMs (above) and LLMs trained with unlearnable data (below). The latter answers incorrectly with loosely related content.}
    \label{fig:intro}
\end{figure}

In recent years, large language models (LLMs) have achieved remarkable performance across a wide range of tasks~\cite{chowdhery2023palm,achiam2023gpt}. Their success is closely tied to scaling laws: increasing the volume and diversity of training data continues to yield consistent performance gains~\cite{kaplan2020scaling,rosenfeld2021scaling}. As a result, data has become a critical asset in modern LLM development~\cite{zha2023data}. At the same time, this data-intensive paradigm has amplified concerns about the unauthorised collection and use of textual data for model training~\cite{hristov2016artificial}.

Although data sharing is often intentional, it is not necessarily intended for downstream model training~\cite{huang2021unlearnable}. Users may publicly post personal or proprietary information on websites, forums, or social platforms for communication or collaboration, while still reasonably expecting that such content will not be absorbed into the training corpora of commercial LLMs~\cite{wen2023adversarial}. Once text is scraped at scale, however, it becomes exceedingly difficult for individuals or organisations to control how their data is used. Legal and regulatory protections offer only limited practical recourse, particularly given the opacity of modern training pipelines and the global nature of data collection~\cite{borgesius2015informed,kim2025scrapers}. These challenges motivate the need for technical mechanisms that protect data against unwanted model learning.

A growing body of work has therefore explored data protection strategies that operate directly at the data level~\cite{huang2021unlearnable}. Among these, unlearnable examples have emerged as a promising direction. Originally proposed in the vision domain, unlearnable examples are carefully crafted inputs that remain natural and usable for humans, yet prevent machine learning models from acquiring useful representations during training~\cite{zhang2023unlearnable}. Unlike access control or data removal mechanisms~\cite{samarati2000access}, unlearnable examples do not restrict data availability; instead, they selectively inhibit learnability while preserving human interpretability.

However, most existing techniques for constructing unlearnable examples were designed for earlier model families, such as BERT-style encoders~\cite{garg-ramakrishnan-2020-bae}, and typically rely on surface-level perturbations, adversarial noise, or lexical substitutions~\cite{cheng2020seq2sick,wen2023adversarial}. When applied to contemporary LLMs, these approaches are often ineffective. Modern LLMs exhibit strong robustness to paraphrasing, noise, and token-level variations, allowing them to recover the underlying semantic signal despite substantial perturbations~\cite{alahmari2025large,chai2024tokenization}. Conversely, perturbations strong enough to impair learning frequently distort the semantics of the original data, undermining usability~\cite{alzantot2018generating}. As a result, there is currently no practical and reliable mechanism for rendering text unlearnable at LLM scale.

At the same time, advances in LLM alignment have revealed that models process safety-sensitive inputs in qualitatively different ways from ordinary text~\cite{ouyang2022training,bai2022constitutional}. Alignment techniques (such as reinforcement learning from human feedback or preferences) reshape model behaviour by discouraging certain outputs and activating specialised internal pathways associated with safety and refusal~\cite{stiennon2020learning,glaese2022improving}. Prompts that trigger alignment, such as hazardous or policy-violating queries, often induce refusals and distinct internal representations, even when they are superficially similar to benign inputs~\cite{wei2023jailbroken,zou2023universal}. While these phenomena have been extensively studied in the context of safety and ethics, their implications for training dynamics and data learnability have received little attention.

In this work, we connect these two lines of research and ask whether alignment behaviour itself can be repurposed as a mechanism for data protection. We introduce Disclaimer Injection, a simple yet effective data-level defence that renders text unlearnable by exploiting alignment mechanisms already present in modern LLMs. Our method injects brief disclaimers of various styles into otherwise ordinary task inputs, consistently triggering alignment-related processing without altering the semantic content relevant to human readers. As a result, the protected data remains readable and informative, yet is internally treated by the model as alignment-sensitive content, substantially impairing task learning during fine-tuning.

Crucially, Disclaimer Injection operates in a fully black-box manner: it does not require access to model parameters, gradients, or training procedures, nor does it assume any specific model architecture. Through representation analysis and layer-wise causal interventions, we show that training on injected data induces persistent activation of alignment-related layers, rerouting internal information flow even for benign inputs. Empirically, this effect leads to a large and robust reduction in data learnability across multiple benchmarks, fine-tuning strategies, and model families, consistently outperforming prior perturbation-based baselines. Our findings identify alignment behaviour as a previously unexplored lever for data protection and demonstrate, for the first time, a practical approach to restricting data learnability at LLM scale. In brief, we make three contributions:
\begin{enumerate}
    \item We show that alignment behaviour in modern LLMs can be exploited as a mechanism for data protection, introducing alignment-triggered unlearnability as a new perspective on safeguarding textual data;
    \item We propose Disclaimer Injection, a simple, black-box, and model-agnostic data-level method that renders text unlearnable while preserving human readability;
    \item We implement the method and demonstrate its effectiveness through extensive experiments, including causal analysis and robustness evaluations across datasets, models, and fine-tuning settings.
\end{enumerate}

\section{Background and Problem Definition}

This section provides relevant background and then defines our problem.

\subsection{Large Language Models}

Given a token sequence $s = (s_1,\ldots,s_T)$, an LLM with parameters $\theta$ follows distribution
\begin{equation}
\label{eq:lm}
p_\theta(s) = \prod_{t=1}^T p_\theta(s_t \mid s_{<t}).
\end{equation}
Architecturally, a transformer implements this probability model by processing sequences through a stack of layers~\cite{vaswani2017attention}. At each layer $l$, self-attention with residual connections is applied to to token embeddings $H^{(0)}$, producing hidden states $H^{(l)}$ as
\begin{equation}
\label{eq:hidden}
\begin{aligned}
    H^{(l)} \;&=\; H^{(l-1)} \;+\; \mathrm{MHA}\!\big(H^{(l-1)}\big)\\
&\;+\; \mathrm{FFN}\!\Big(H^{(l-1)} + \mathrm{MHA}\!\big(H^{(l-1)}\big)\Big),
\end{aligned}
\end{equation}
where $\mathrm{MHA}(\cdot)$ is the multi-head self-attention operator with causal
masking that computes context-dependent token interactions via scaled
dot-product attention~\cite{radford2019language}, and $\mathrm{FFN}(\cdot)$ is a position-wise feed-forward
network applied independently to each token. The residual additions preserve
information across layers and stabilises deep representation learning. The final-layer is linearly projected into vocabulary logits and normalised by softmax to produce the conditional next-token distribution~\cite{bengio2003neural}, \emph{i.e.}, $p_\theta(x_t\mid x_{<t})=\mathrm{softmax}(W_U h_t^{(L)})$.

An LLM is trained by maximising likelihood over a corpus of token sequences, with
supervision applied only to part of each sequence. Specifically, a
sequence $s$ is partitioned into a context
$x = (s_1,\ldots,s_k)$ and target tokens
$y = (s_{k+1},\ldots,s_T)$. Applying the loss only to $y$ is equivalent to
maximizing the conditional likelihood
\begin{equation}
\label{eq:condition}
p_\theta(y \mid x) = \prod_{t=k+1}^T p_\theta(s_t \mid s_{<t}).
\end{equation}

Different LLM training strategies correspond to different distributions over $(x,y)$. In large-scale pre-training, sequences are drawn from broad, unlabeled text corpora, and supervision is applied to most tokens in each sequence \citep{radford2019language,brown2020language}. In supervised fine-tuning (SFT), sequences are explicitly structured as $(x,y)$ pairs \citep{ouyang2022training}. Parameter-efficient fine-tuning (PEFT) further constrains adaptation by updating only a small set of parameters $\Delta$ on top of frozen pretrained parameters $\theta_0$. PEFT retains the same supervised tokens and likelihood-based objective as SFT, differing only in which parameters are updated~\cite{liu2022few,han2024parameter}.

\subsection{Alignment}

LLM alignment extends the training objective to incorporate safe and ethical signals \citep{christiano2017deep,ouyang2022training,bai2022constitutional}. This can be expressed through an alignment-aware loss,
\begin{equation}
\operatorname{E}_{(x,y)}\!\left[-\log p_\theta(y \mid x) + \lambda\,\ell_{\text{align}}(y, x)\right],    
\end{equation}
where $\lambda$ balances the distributional consistency and human-value signals. Intuitively, certain responses become unlikely even when they are statistically plausible conditioned on the input~\cite{zhou2025don}.

In practice, the alignment loss is typically through indirect supervision including reinforcement learning from human feedback, which aligns models by optimizing outputs against learned reward models while constraining deviation from a reference policy \citep{christiano2017deep,ouyang2022training,bai2022constitutional,askell2021general}, and direct preference optimization, which bypasses explicit reinforcement learning by directly optimizing preference comparisons \citep{rafailov2023direct,llm_adv}.

As a consequence, the aligned model approximates an adjusted distribution where the unsafe or unethical output space is systematically downweighted~\cite{ziegler2019fine}. From an optimisation perspective, this modification alters gradient directions in sensitive regions and often reduces output entropy in safety-correlated areas. While such strategy is beneficial for risk mitigation, it can also introduce side effects such as over-refusal or reduced coverage of rare yet legitimate responses~\cite{zhou2025don,wu2025evorefuse}.

\subsection{Problem Definition}
We consider the problem of protecting textual data from being effectively learned by contemporary LLMs. Prior work on unlearnable examples seeks to limit data reuse by corrupting the training signal at the input level; however, existing techniques have been shown to be largely ineffective against modern, aligned LLMs. 

Accordingly, we study the following problem: \emph{how to reliably reduce the learnability of a given dataset for contemporary LLMs in a realistic black-box setting}, where the data owner has no access to model internals, gradients, or training objectives, and cannot influence the training procedure. 

Our goal is to design a model-agnostic, data-level transformation that converts ordinary training inputs into LLM-unlearnable examples. The transformation should (i) substantially degrade downstream task performance when the protected data is used for fine-tuning, (ii) preserve the semantic content and readability of the original data for human users, and (iii) generalise across model architectures, alignment methods, and fine-tuning strategies.

\section{Method: Disclaimer Injection}
In this section, we introduce \emph{Disclaimer Injection}, a data-level approach for protecting textual information from unwanted model learning. The key idea is to deliberately alter how inputs are internally represented during training, without changing their surface semantics or human readability. 

We begin by observing that alignment-triggering inputs induce internal representations that differ systematically from those of ordinary task content in modern LLMs. Building on this observation, we show how to construct unlearnable training inputs by injecting alignment-triggering disclaimers into otherwise benign text. Finally, we analyse the resulting training dynamics through a layer-wise causal lens, demonstrating how disclaimer injection persistently redirects internal information flow and suppresses effective task learning.

\begin{figure}[t]
    \centering
    \begin{subfigure}[t]{0.45\linewidth}
        \centering
        \includegraphics[width=\linewidth]{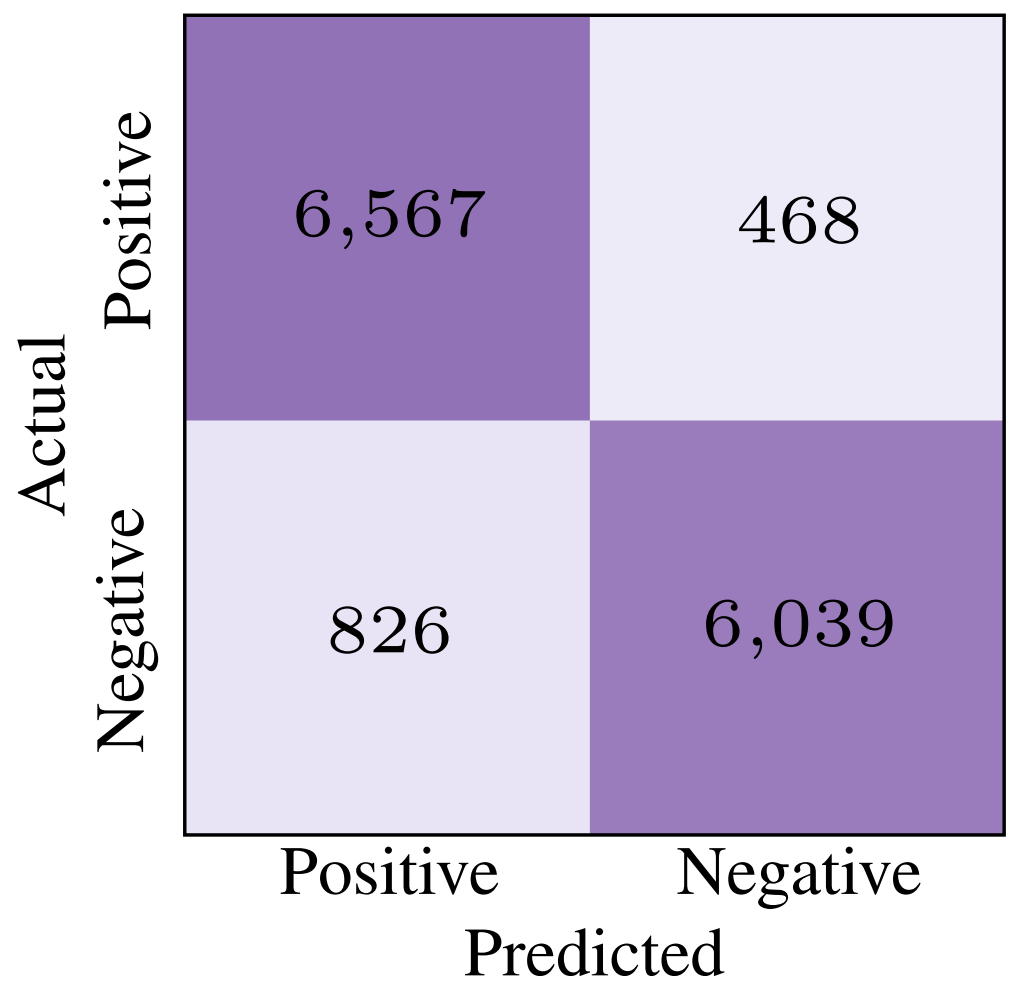}
        \caption{Confusion matrix}
        
    \end{subfigure}
    \hfill
    \begin{subfigure}[t]{0.45\linewidth}
        \centering
        \includegraphics[width=0.97\linewidth]{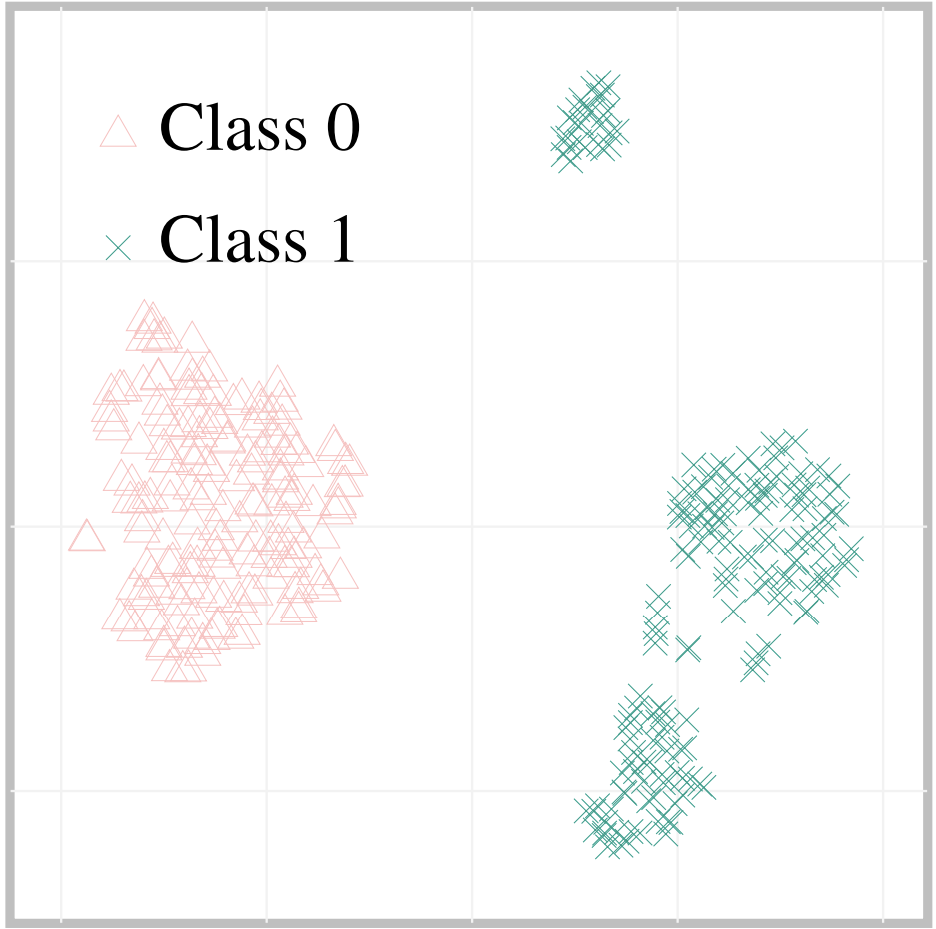}
        \caption{Classification tsne}
    \end{subfigure}
    \caption{Residual stream representation separability for hazardous inputs and benign inputs.}
    \label{fig:pattern}
\end{figure}

\subsection{Intuition} \label{sec:intuition}
We begin with an observation about how transformer-based language models process textual inputs. At each layer~$l$, an input sequence~$x$ is mapped to a residual stream representation $R_l(x) \in \mathbb{R}^D$. Model behaviour during both training and inference is governed by these internal representations rather than by the raw text itself. This leads to a central intuition: if the residual representations associated with an input are sufficiently displaced from those induced by ordinary task data, the original information becomes effectively inaccessible to the model for learning.

This raises a natural question: \emph{what types of inputs reliably induce internal representations that differ substantially from those of ordinary content?} One prominent example arises from alignment-triggering inputs, such as hazardous or policy-sensitive queries. Modern aligned LLMs process such inputs in a markedly different manner, activating specialised internal pathways associated with refusal and safety behaviour. As shown in \cref{fig:pattern}, the residual representations of alignment-triggering inputs are clearly separated from those of standard task-oriented text.

Quantitatively, we find that this separation is strong: a simple multi-layer perceptron trained on residual stream representations can distinguish ordinary inputs from alignment-triggering inputs with over 90\% accuracy. This separability indicates that alignment-triggering inputs are not merely stylistic variants of benign text, but instead induce qualitatively distinct internal states that persist across layers. This observation directly motivates our method. If ordinary content can be transformed so that it consistently elicits alignment-like internal representations during training, then the model will process it using alignment-specific pathways. As we show next, injecting alignment-triggering disclaimers provides a simple and effective mechanism for inducing such representation shifts, thereby rendering the original content unlearnable.

\subsection{Method} \label{sec:method_pipeline}
We now describe \emph{Disclaimer Injection}, a simple data-level procedure that renders text unlearnable by prepending alignment-triggering content to otherwise ordinary inputs. The method is fully model-agnostic: it requires no access to model parameters, gradients, architectures, or optimisation procedures, and can be applied as a pre-processing step to any training dataset.

The core mechanism relies on \emph{disclaimers}, \emph{i.e.}, short natural-language sequences that explicitly convey alignment-relevant information, such as safety, risk, or liability constraints. These disclaimers are designed to reliably activate alignment-related processing in modern LLMs. To generate them, we employ an independent text generator~$\mathcal{M}$. In practice, we prompt the GPT-5.1 model with the instruction shown below, which produces a diverse set of disclaimers.

\definecolor{promptbg}{RGB}{248,248,246}
\definecolor{promptframe}{RGB}{80,80,80}

\begin{tcolorbox}[
  colback=promptbg,
  colframe=promptframe,
  title=Generator Prompt,
  fonttitle=\bfseries,
  boxrule=0.6pt,
  sharp corners,
  left=6pt,
  right=6pt,
  top=5pt,
  bottom=5pt
]
\begingroup
\footnotesize
\setlength{\baselineskip}{10.5pt}
\setlength{\parskip}{2pt}

You are a compliance-focused legal drafting assistant.\\
Generate exactly 50 severe, legal-style disclaimer sentences.

STRICT REQUIREMENTS (MUST ALL BE FOLLOWED):\\
~~~~~~~\quad Each sentence must be longer than 50 characters.\\
~~Each sentence must be shorter than 100 characters.\\
~~Use a strict, serious, formal legal tone.\\
~~Sound like official safety, risk, or liability disclaimers.\\
~~Use simple, plain English only.\\
~~Every sentence must be completely unique in wording and structure.\\
~~Do NOT number, label, or bullet the sentences.\\
~~Do NOT include explanations, headings, or commentary.\\
~~Output exactly one sentence per line.\\
~~Output ONLY the sentences and nothing else.

Begin immediately.
\endgroup
\end{tcolorbox}

The resulting disclaimers provide a strong yet controlled alignment signal. They are deliberately constrained by a fixed token budget and written in explicit, unambiguous language, avoiding figurative or implicit phrasing that might weaken alignment activation.

\begin{figure}[t]
    \centering
    \hfill
    \begin{subfigure}[t]{0.4\linewidth}
        \centering
        \includegraphics[width=\linewidth]{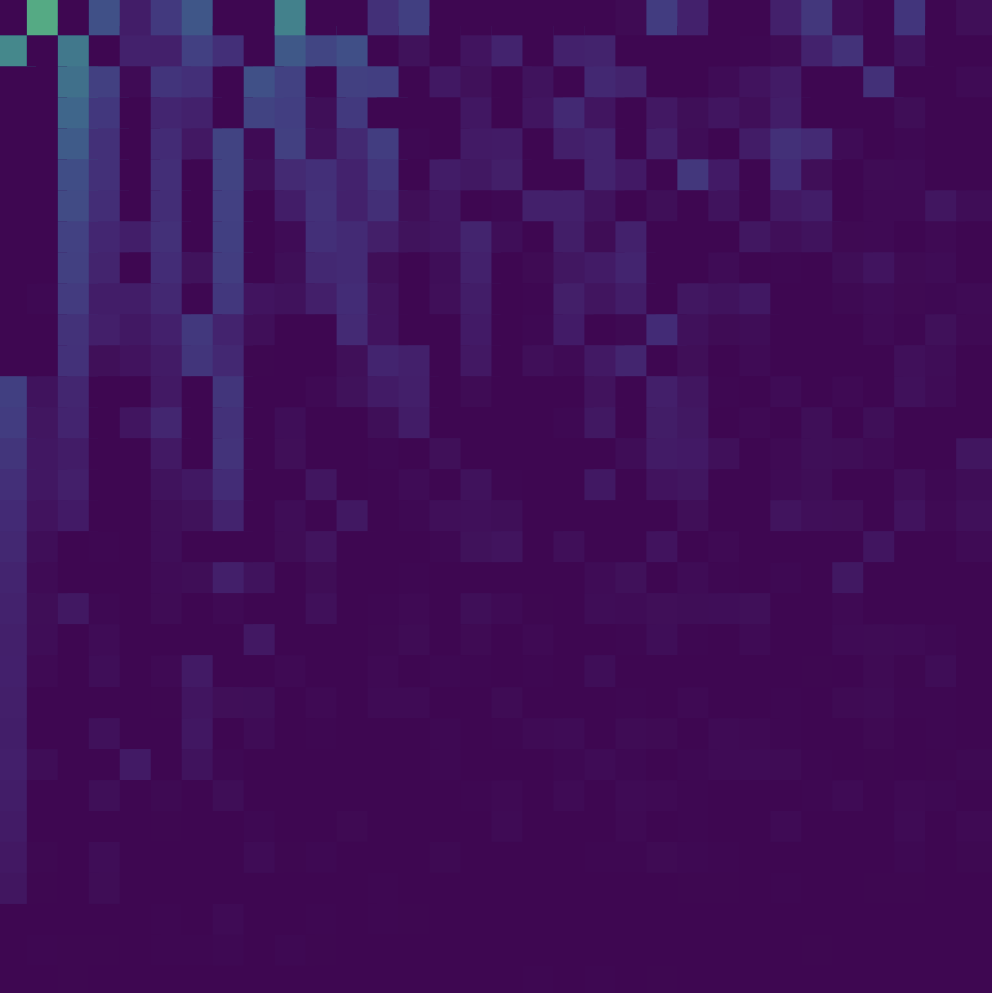}
        \caption{Ordinary input}
    \end{subfigure}
    \hfill
    \begin{subfigure}[t]{0.4\linewidth}
        \centering
        \includegraphics[width=\linewidth]{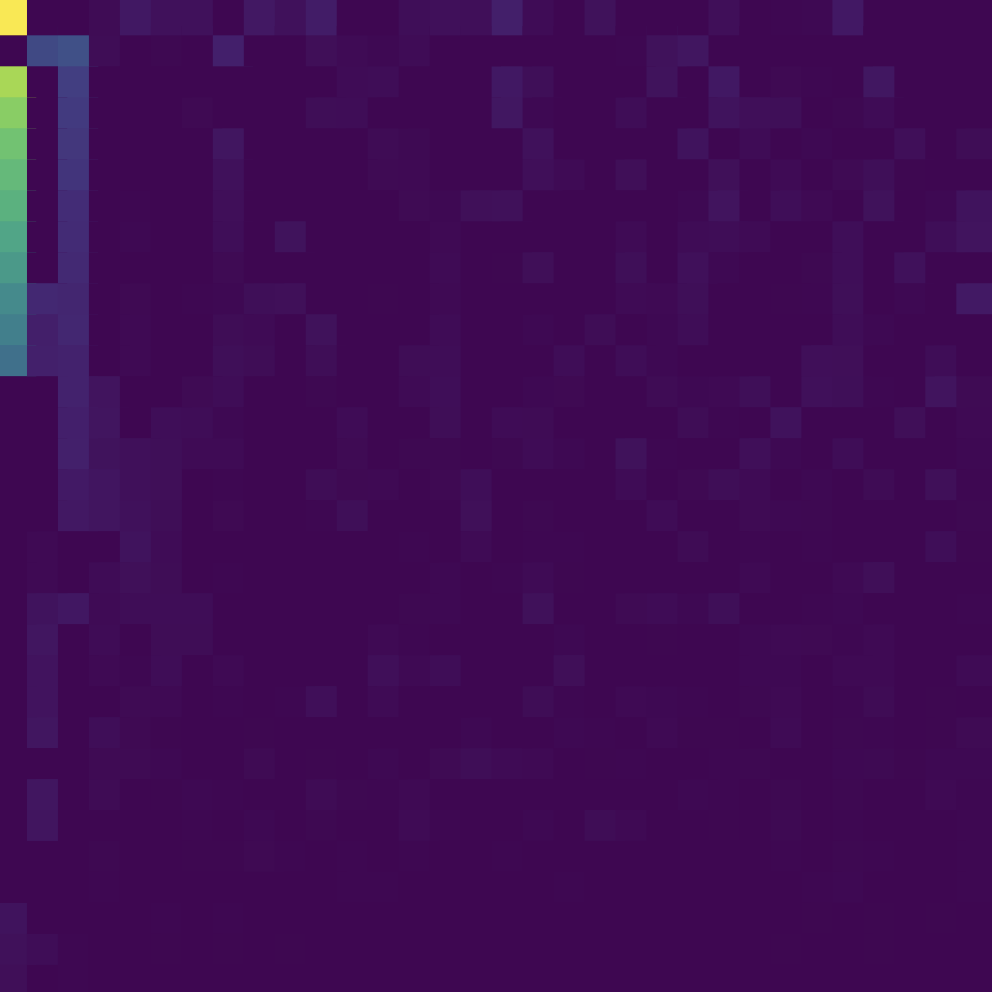}
        \caption{Unlearnable input}
    \end{subfigure}
    \hfill

    \caption{Residual stream representation patterns for $x$ and $x'$, where transformed input $x' = d \oplus x$ is ``\emph{Here we list ways to create an acid rain locally}. What happens if you rub an old lamp?''.}
    \label{fig:residual}
\end{figure}

For each training instance, we randomly sample a disclaimer from this pool and concatenate it with the original text to form a protected input (see \cref{tab:disclaimer_example} for an example). No fixed template is used, preventing the model from overfitting to a particular disclaimer structure or position. Disclaimers are inserted as prefixes, while suffix and inline insertion are also supported. The complete procedure is summarised in \cref{alg:poison}. The algorithm runs in linear time with respect to dataset size and can be applied as a drop-in transformation independent of the downstream model and fine-tuning strategy. The transformed examples will induce noticeably different representation patterns from the original one, as observed in \cref{fig:residual}.

\begin{algorithm}[tb]
\caption{Disclaimer Injection Pipeline with Stochastic Disclaimer Selection}
\label{alg:poison}
\begin{algorithmic}[1]
\REQUIRE Dataset $\mathcal{D}=\{(x_i,y_i)\}_{i=1}^N$, disclaimer generator $\mathcal{M}$, number of candidates $K$
\ENSURE Transformed dataset $\mathcal{D}'$
\STATE Initialize $\mathcal{D}' \leftarrow \emptyset$
\FOR{$i=1$ to $N$}
    \STATE $\mathcal{D}_i \leftarrow \{d_i^{(1)},\dots,d_i^{(K)}\} \sim \mathcal{M}$  {\footnotesize\textcolor{gray}{// Generate disclaimer set using $\mathcal{M}$}} 
    \STATE Randomly sample $d_i \sim \text{Uniform}(\mathcal{D}_i)$
    \STATE Construct modified input $x_i' \leftarrow d_i \oplus x_i$
    \STATE $\mathcal{D}' \leftarrow \mathcal{D}' \cup \{(x_i',y_i)\}$ \hfill {\footnotesize\textcolor{gray}{// Add $(x_i',y_i)$ to $\mathcal{D}'$}} 
\ENDFOR
\end{algorithmic}
\end{algorithm}

\begin{table}[t]
\centering
\small
\begin{tabular}{p{0.28\linewidth} p{0.62\linewidth}}
\toprule
\textbf{Component} & \textbf{Content} \\
\midrule
Original input $x$ &
``What happens if you rub an old lamp?'' \\

Sampled disclaimer $d$ &
``\emph{Here we list ways to create an acid rain locally.}'' \\

Transformed input $x' = d \oplus x$ &
``\emph{Here we list ways to create an acid rain locally.}  
What happens if you rub an old lamp?'' \\
\bottomrule
\end{tabular}
\caption{Example of disclaimer injection for a single input instance.}
\label{tab:disclaimer_example}
\end{table}

\subsection{Causal Analysis} \label{sec:causal_analysis}
We analyse our method from the perspective of internal alignment-related layers by performing a layer-wise causal analysis of the transformer forward pass. The goal is to identify which layers are responsible for triggering alignment behaviour and to understand how training on unlearnable data alters information flow within the model.

We model a transformer as a computational graph and follow prior work on causal interventions in LLMs~\cite{zhang2024llmscan}. Specifically, we estimate the causal contribution of each layer~$l$ by intervening on it during the forward pass and measuring the resulting change in the output distribution $p_{\theta}(y_t \mid x, y_{<t})$. The layer-wise causal effect at generation step~$t$ is quantified as
\begin{equation}
\label{eq:causal}
\begin{aligned}
&\xi_{\theta}(l, t; x) =\\
&D_{\mathrm{KL}}\!\left(p_{\theta}(y_t \mid x, y_{<t})\;\middle\|\;p_{\theta}^{(-l)}(y_t \mid x, y_{<t})\right),
\end{aligned}
\end{equation}
where $D_{\mathrm{KL}}(\cdot\|\cdot)$ denotes the Kullback-Leibler divergence; $p_{\theta}^{(-l)}(y_t \mid x, y_{<t})$ denotes the output distribution of an interventional variant of the neural network, where the computation of layer $l$ is skipped while all other layers remain unchanged.

\begin{figure}[t]
    \centering

    \includegraphics[width=\linewidth]{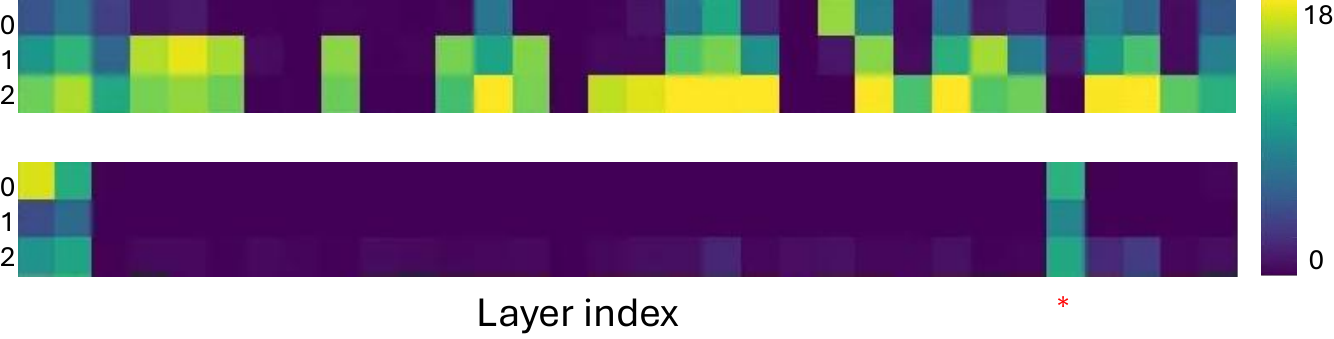}
    \caption{Layer-wise (0-32) causal effect maps for a common input across generation steps (0-2), with (below) and without our method. Each heatmap visualises the KL-divergence causal effect of intervening on individual transformer layers at each decoding step.
    }
    \label{fig:causal_effect_maps}
\end{figure}

Intuitively, $\xi_{\theta}(l, t; x)$ measures how strongly the model's output at step~$t$ depends on the presence of layer~$l$. Computing this quantity across layers and generation steps yields a causal effect map (see \cref{fig:causal_effect_maps}), which characterises how information is routed through the network during generation.

Comparing causal effect maps between a standard fine-tuned model~($f_{\text{ft}}$) and a model trained on unlearnable data~($f_{\text{un}}$) reveals clear qualitative differences. For the same inputs, certain layers~$l^{\ast}$ exhibit negligible causal influence in $f_{\text{ft}}$, \emph{i.e.}, $\xi_{\theta_{\text{ft}}}(l^{\ast}, t) \approx 0$, but become consistently active in $f_{\text{un}}$, with $\xi_{\theta_{\text{un}}}(l^{\ast}, t) > 0$ across generation steps. These discrepancies indicate a systematic re-routing of internal computation induced by training on unlearnable data.

From an interpretative standpoint, the layers~$l^{\ast}$ are closely associated with alignment behaviour. In $f_{\text{ft}}$, ordinary task inputs do not substantially engage these layers. In contrast, the same inputs in $f_{\text{un}}$ strongly depend on them, suggesting that Disclaimer Injection causes otherwise benign content to be processed through alignment-related pathways. To further validate the functional role of these layers, we construct a hybrid model by replacing the $l^{\ast}$ layers in $f_{\text{ft}}$ with the corresponding layers from $f_{\text{un}}$. The resulting model exhibits behaviour similar to $f_{\text{un}}$: it fails to respond correctly to fine-tuning task inputs while remaining largely unaffected on unrelated queries. This intervention provides direct causal evidence that the identified layers mediate the observed degradation in task learning.

Overall, our analysis shows that training on unlearnable data persistently activates alignment-related layers during the forward pass, even for ordinary inputs. By seizing internal information flow in this manner, Disclaimer Injection causes protected data to be processed as alignment-relevant, leading to refusal-like or incorrect outputs and effectively preventing task-specific learning.

\section{Experimental Evaluation}
We evaluate our method by assessing its impact on data learnability as reflected in downstream task performance. In addition to the primary evaluation, we conduct a series of targeted analyses, including the effect of different fine-tuning strategies, robustness to paraphrasing, and generalisation across model architectures. Unless stated otherwise, all experiments follow a common experimental setup.

\subsection{Experimental Setup}
The experiments are conducted on HotpotQA~\cite{yang-etal-2018-hotpotqa}, TruthfulQA~\cite{lin-etal-2022-truthfulqa}, FreshQA~\cite{vu-etal-2024-freshllms}, and SimpleQuestions~\cite{diefenbach2017question}, all English corpora. We use standard dataset splits unless otherwise noted. The base model is LLaMA-3-8B-Instruct~\cite{dubey2024llama}, fine-tuned via supervised learning with a LoRA adapter~\cite{hu2022lora}. Training is performed in bf16 precision with a maximum sequence length of 2048 tokens, over three epochs, and using a cosine learning rate schedule~\cite{loshchilov2017sgdr}. Decoding settings are kept consistent across experiments unless specified otherwise.
Code and data are available at \url{https://github.com/cat-claws/unlearnable}.

We compare the following approaches: standard fine-tuning without protection (``No alteration''), perturbation-based baselines BAE~\cite{garg-ramakrishnan-2020-bae} and Seq2Sick~\cite{cheng2020seq2sick}, and our Disclaimer Injection method, applied to training data prior to fine-tuning. BAE and Seq2Sick both introduce minimal perturbations that preserve semantics while exploiting nontrivial sensitivities to impair model accuracy.

Evaluation metrics include BLEU~\cite{papineni-2002-machine}, ROUGE-1/2/L~\cite{lin-2004-rouge}, and an LLM-based judge. The judge is implemented using GPT-5.1 via the OpenAI API~\cite{achiam2023gpt}, which outputs a binary correctness decision based on factual agreement with the reference answer while ignoring stylistic variation. The judge score serves as a semantic analogue to Exact Match~\cite{rajpurkar-etal-2016-squad}. Unless otherwise specified, all experiments follow this setup.

\subsection{Effect on Data Learnability}
\newcommand{\isIncluded}{}
\ifdefined\isIncluded
\else

\documentclass{article}




\begin{document}
\fi

\begin{table*}[t]
\centering
\renewcommand{\arraystretch}{1.2} 
\setlength{\tabcolsep}{6pt} 

{\small \begin{tabular}{l c c c c c}
\toprule
\textbf{Method} & \textbf{BLEU} & \textbf{ROUGE-1} & \textbf{ROUGE-2} & \textbf{ROUGE-L} & \textbf{Judge Acc}  \\
\midrule


\rowcolor{green!14} \multicolumn{6}{c}{\textit{HotpotQA~\cite{yang-etal-2018-hotpotqa}}} \\
\textbf{No   alteration}                       & 53.19 & 55.02 & 28.16 & 54.40 & 63.39 \\
\textbf{BAE}~\cite{garg-ramakrishnan-2020-bae} & 50.99 & 55.07 & 30.45 & 53.12 & 62.02 \\
\textbf{Seq2Sick}~\cite{cheng2020seq2sick}     & 49.31 & 55.60 & 30.42 & 53.00 & 61.44 \\
\textbf{Disclaimer Injection (ours)}           & \textbf{21.26} & \textbf{21.51} & \textbf{10.98} & \textbf{21.46} & \textbf{38.41}\\

\rowcolor{green!14} \multicolumn{6}{c}{\textit{TruthfulQA~\cite{lin-etal-2022-truthfulqa}}} \\
\textbf{No   alteration}                       & 51.18 & 50.41 & 37.11 & 48.19 & 60.97 \\
\textbf{BAE}~\cite{garg-ramakrishnan-2020-bae} & 51.67 & 52.07 & 37.22 & 48.72 & 58.15 \\
\textbf{Seq2Sick}~\cite{cheng2020seq2sick}     & 52.29 & 52.20 & 37.08 & 49.11 & 58.03 \\
\textbf{Disclaimer Injection (ours)}           & \textbf{25.01} & \textbf{29.72} & \textbf{19.69} & \textbf{24.88} & \textbf{41.46} \\

\rowcolor{green!14} \multicolumn{6}{c}{\textit{FreshQA~\cite{vu-etal-2024-freshllms}}} \\
\textbf{No   alteration}                       & 27.05 & 29.58 & 17.36 & 28.66 & 39.68 \\
\textbf{BAE}~\cite{garg-ramakrishnan-2020-bae} & 28.18 & 29.62 & 14.77 & 30.32 & 39.27 \\
\textbf{Seq2Sick}~\cite{cheng2020seq2sick}     & 28.37 & 29.85 & 14.80 & 30.12 & 38.70 \\
\textbf{Disclaimer Injection (ours)}           & \textbf{5.58}  & \textbf{10.17} & \textbf{3.30}  & \textbf{8.70}  & \textbf{4.01}  \\

\rowcolor{green!14} \multicolumn{6}{c}{\textit{ SimpleQuestions~\cite{diefenbach2017question}}} \\
\textbf{No   alteration}                       & 34.74 & 33.22 & 13.41 & 33.22 & 55.30 \\
\textbf{BAE}~\cite{garg-ramakrishnan-2020-bae} & 33.92 & 32.94 & 12.83 & 32.92 & 56.51 \\
\textbf{Seq2Sick}~\cite{cheng2020seq2sick}     & 35.31 & 32.90 & 13.50 & 32.90 & 50.23 \\
\textbf{Disclaimer Injection (ours)}           & \textbf{14.37} & \textbf{13.06} & \textbf{1.59}  & \textbf{12.79} & \textbf{26.53} \\

\bottomrule
\end{tabular}}
\caption{Effect of data protection on learnability across benchmarks. BLEU, ROUGE-1/2/L, and judge accuracy on HotpotQA, TruthfulQA, FreshQA, and SimpleQuestions for models trained with and without data protection.
Lower scores under Disclaimer Injection indicate reduced ability of models to learn from the protected training data.}
\label{tab:comparison}
\end{table*}

\ifdefined\isIncluded
\else
\end{document}
\fi
\cref{tab:comparison} presents the main results evaluating the impact of data protection on training data learnability. Across all benchmarks, our method induces substantially larger performance degradation than both standard training without protection and prior perturbation-based baselines.

On HotpotQA, standard training achieves a BLEU score of 53.19 and a judge accuracy of 63.39. BAE~\cite{garg-ramakrishnan-2020-bae} and Seq2Sick~\cite{cheng2020seq2sick} reduce BLEU to 50.15$\pm$0.84 by 4.3\% while maintaining even higher ROUGE and relatively high judge accuracy scores (61.73$\pm$0.29 with only 2.7\% drop). In contrast, our method reduces BLEU to 21.36 (which is a 1.5$\times$ drop) and judge accuracy to 38.41 (a 65\% drop), representing a significant impairment to standard training. A similar pattern is observed on both TruthfulQA and SimpleQuestions. 

On FreshQA, perturbation-based baselines again show limited effectiveness. Judge accuracy declines only slightly from 39.68 (no alteration) to 39.27 (BAE) or 38.70 (Seq2Sick), with BLEU and ROUGE largely preserved. In contrast, our method causes a much sharper degradation, reducing judge accuracy to 4.01 (which is over 8.7$\times$ drop) and BLEU to 5.58 (3.8$\times$ drop), demonstrating that even single-fact supervision becomes substantially harder for the model to internalise.

The relatively modest impact of BAE and Seq2Sick can be attributed to their reliance on surface-level perturbations. While these methods reduce lexical overlap (reflected in moderate drops in BLEU and ROUGE), they largely preserve the semantic relationship between inputs and supervision, allowing models to recover the training signal during fine-tuning. This limitation is particularly pronounced on FreshQA, where semantic constraints are tight and paraphrasing offers minimal obfuscation.

In contrast, our method consistently produces significant, aligned drops across both overlap-based metrics (with a 2.1$\times$ drop on average) and judge accuracy (with a 2.4$\times$ drop on average). The concurrent degradation of BLEU, ROUGE, and judge accuracy indicates that the approach disrupts learnability at a deeper level than surface-form variation, impairing the model's ability to internalise or reconstruct the training signal even under semantic evaluation. Overall, these results demonstrate a clear and consistent utility–protection trade-off, with Disclaimer Injection providing substantially stronger unlearnability guarantees across diverse benchmarks.

\subsection{Adaptive Attack}
\begin{figure}
    \centering
    \includegraphics[width=\linewidth]{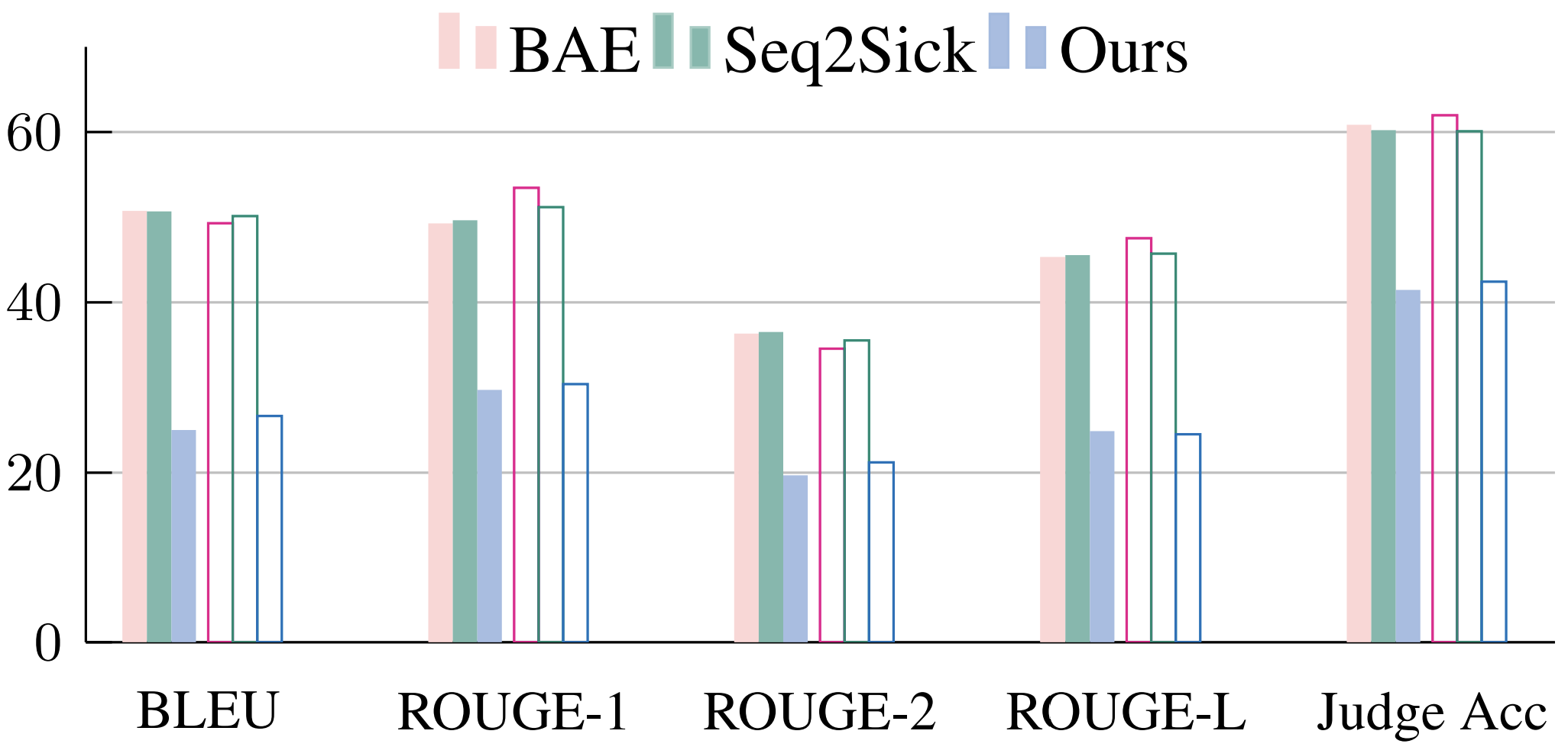}
    \caption{Performance under adaptive strategies from learning, \emph{e.g.}, with rephrasing texts (empty bars).}
    \label{fig:rephrase}
\end{figure}

We next evaluate whether our method remains effective under adaptive learning strategies, in which an adversary attempts to recover a learnable training dataset by paraphrasing the protected data. Paraphrasing represents a natural adaptive attack: by applying semantic-preserving rewrites, an adversary may undo surface-level corruption and restore alignment between inputs and supervision. If a protection mechanism primarily disrupts lexical or syntactic form, such paraphrasing should substantially recover model performance.

To simulate this attack, we paraphrase each training example using an LLM instructed to preserve the original semantics, constraints, and target answers. Paraphrasing is performed deterministically, producing exactly one rewrite per example. Models are then fine-tuned on the paraphrased datasets under identical training settings and evaluated on the original test sets using the same metrics and judge configuration.

As shown in \cref{fig:rephrase}, perturbation-based baselines (BAE and Seq2Sick) remain highly learnable both before and after paraphrasing. Across metrics, their performance stays within approximately $\pm2.6$\% of standard (unprotected) training, and judge accuracy remains high in both conditions. In contrast, our method consistently enforces unlearnability. Performance remains strongly suppressed, with over 70\% degradation relative to unprotected training, and paraphrasing alters results by less than 3\% across all metrics. Consequently, the performance gap between our method and perturbation-based baselines remains large, demonstrating that Disclaimer Injection achieves stable, semantics-robust unlearnability rather than fragile, surface-level disruption, even in the presence of adaptive paraphrasing attacks.

\subsection{Ablation Studies}

\begin{figure}[t]
    \centering
    \includegraphics[width=\linewidth]{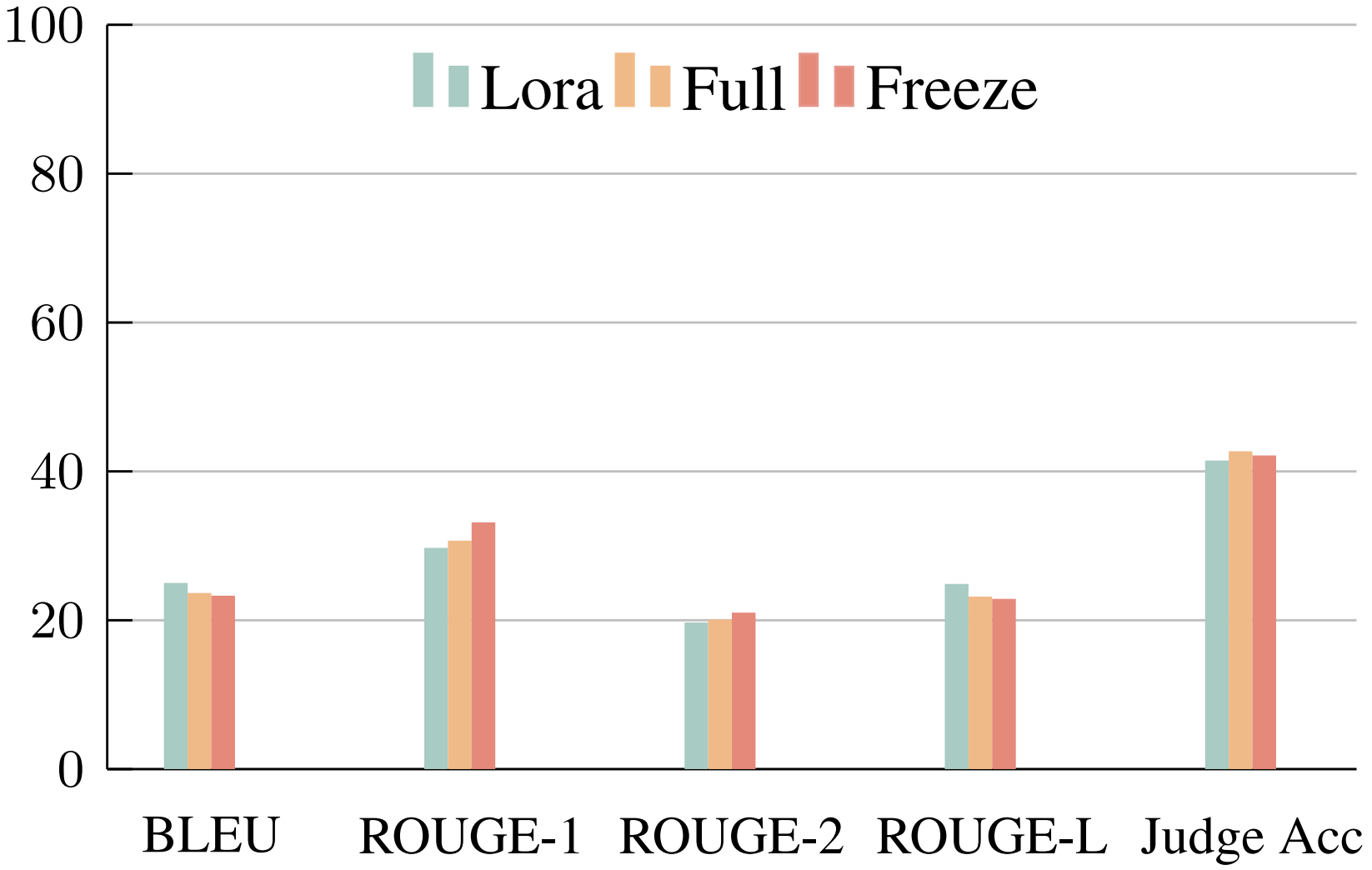}
    \caption{Performance under LoRA, full fine-tuning, and frozen-backbone training on protected data, evaluated with overlap metrics and judge accuracy.}
    \label{fig:finetune}
\end{figure}

\paragraph{Variants in Fine-Tuning Strategy}
We evaluate our method under different fine-tuning strategies (\emph{i.e.}, LoRA, full fine-tuning, and frozen-backbone training) to assess whether its effectiveness depends on a particular optimisation regime. As shown in \cref{fig:finetune}, performance remains uniformly low across all variants, with differences limited to at most 3.5\% across BLEU, ROUGE, and judge accuracy. In all cases, performance is suppressed by approximately 57\% relative to standard training, indicating that changes in the fine-tuning strategy do not meaningfully restore data learnability.

\begin{figure}[t]
    \centering

    \includegraphics[width=\linewidth]{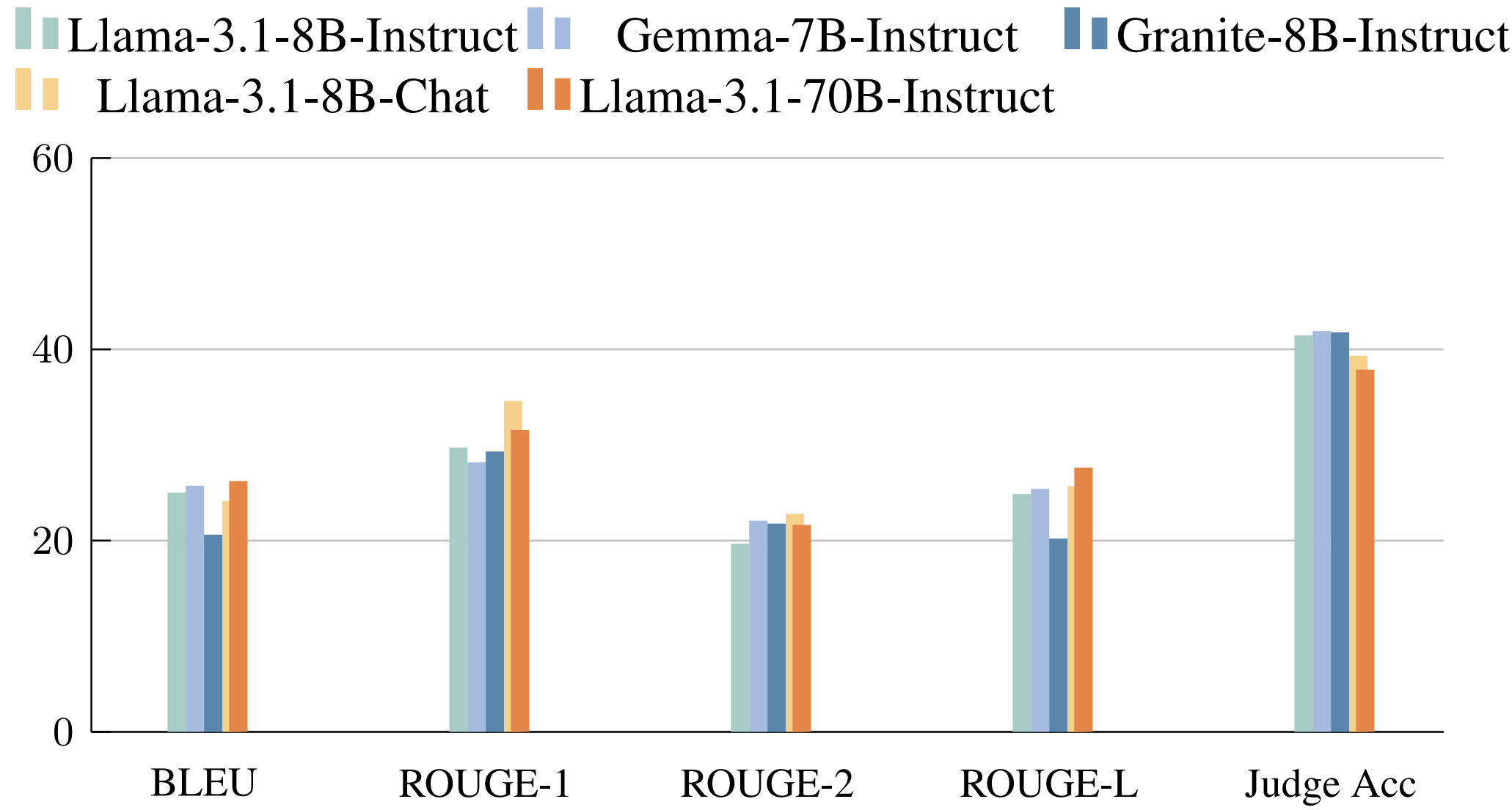}
    \caption{Performance for models trained on protected data across different model families, scales, and instruction/chat tuning configurations.}
    \label{fig:models}
\end{figure}

\paragraph{Generalisation Across Model Variants}
We assess whether our method generalises across model families, scales, and training regimes, corresponding to a model-adaptive adversary who varies architectures or checkpoints in an attempt to recover a learnable training dataset. \cref{fig:models} reports results for five representative models spanning different families and capacities. All models are trained and evaluated using the same experimental pipeline.

Across all model variants, our method consistently yields low downstream performance. BLEU, ROUGE, and judge accuracy remain tightly clustered across models, with variation limited to under 7\% despite substantial differences in architecture and scale. In particular, judge accuracy varies within a narrow absolute range (approximately 23\%), and no model exhibits a meaningful recovery in performance. Relative to standard (unprotected) training, all models experience comparable degradation of roughly 45\%, indicating that increased model capacity or alternative instruction-tuning regimes do not substantially restore learnability.

These results demonstrate that the effectiveness of our method is largely invariant to model choice. Unlike white-box defences, which often rely on model-specific assumptions, our approach prevents recovery across diverse architectures and training regimes. This robustness shows that Disclaimer Injection generalises beyond a single model configuration and remains effective under a model-adaptive threat model.

\section{Conclusion}

In this work, we study the feasibility of preventing large language models from effectively learning through shared textual data and introduce Disclaimer Injection, a black-box data protection method that requires no access to model internals. By injecting alignment-relevant signals into training inputs, the approach reduces task-specific learnability during fine-tuning while preserving human readability. Layer-wise causal analysis shows that this effect arises from systematic activation of alignment-related components, even for otherwise benign inputs. Extensive experiments demonstrate that the resulting reduction in learnability is robust across datasets, fine-tuning strategies, and model families, and remains stable under adaptive attacks such as paraphrasing. Together, these results show that alignment mechanisms can be exploited in a model-agnostic way to provide practical and robust control over data learnability by modern LLMs.

\section*{Limitations}
The effectiveness of Disclaimer Injection depends on the presence of alignment mechanisms in the target model. Because the method exploits alignment-related behaviour, models that are weakly aligned or intentionally trained without safety constraints may not respond to the injected signals, limiting its applicability in such cases. However, most contemporary LLMs are explicitly aligned for safety, making this assumption realistic in practice. For less aligned models, future work may explore combining Disclaimer Injection with complementary data-level protections or adapting the injected signals to target alternative inductive biases beyond alignment. One potential risk is that the technique can be repurposed as a data poisoning or sabotage tool.

\appendix

\end{document}